\documentclass[runningheads]{llncs}
\usepackage[T1]{fontenc}
\usepackage{graphicx}
\usepackage{array}
\usepackage{tabularx}
\usepackage{booktabs}
\usepackage{multirow}
\usepackage{subcaption}
\usepackage{float}

\begin{document}
\title{Towards Zero-Shot Domain Generalization for ID Cards Presentation Attack Detection}
\titlerunning{Towards Few-Shot Domain Generalization for ID Cards PAD}
%
\author{Mario Nieto-Hidalgo\inst{1}\orcidID{0000-0003-0623-6455} \and
Juan M. Espin\inst{1}\orcidID{0000-0001-6521-7890} \and
Juan E. Tapia\inst{2}\orcidID{0000-0001-9159-4075}}
\authorrunning{Nieto-Hidalgo et al.}
%
\institute{Facephi, R\&D Department, Spain. \email{\{marionieto,jmespin\}@facephi.com}\\
\and
da/sec-Biometrics and Internet Security Research Group. \\ 
Hochschule Darmstadt, Germany. \email{juan.tapia-farias@h-da.de}}
\maketitle              
\begin{abstract}
Presentation-Attack Detection (PAD) for national ID cards is limited by the lack of publicly available genuine samples, making it difficult for systems to generalize across countries. This paper introduces two main innovations: (1) a Prototypical Network head using an EfficientNet-V2-b0 backbone that requires only four genuine samples per class to create reliable prototypes; and (2) an episodic training regime that keeps PAD classes fixed while varying the card domain, allowing the network to learn universal attack cues.

Evaluated on a large multi-country dataset and the public DLC-2021 benchmark, this method achieves an average Equal Error Rate of around 9\%, outperforming conventional softmax and CLIP zero-shot baselines even with data from a single source country. This approach provides accurate, privacy-preserving PAD while minimizing data collection, facilitating scalable cross-jurisdictional remote onboarding.

\end{abstract}
\section{Introduction}
\label{sec:introduction}

Growing concerns about biometric companies in remote verification processes have increased the popularity of fully online onboarding systems. The popularity of these systems is mainly due to their user-friendly enrollment and the widespread adoption of smartphones. Users can enroll directly from home by taking a selfie or a photo of their ID card, without needing to go to a store or an office. These factors increase the number of users willing to enroll. However, this system carries a high risk of fraud, as users might attempt to exploit it. The easiest way to fool these systems is with Presentation Attacks (PAs), in which the fraudster impersonates another subject using a printed photo or a screen replay and presents it to the capture device. These systems prevent this type of fraud through PAD algorithms. Today, most solutions are based on algorithms, usually Deep Learning (DL) networks, trained on large amounts of images to distinguish high-frequency details in PAs, such as the Moiré patterns of screen-replay attacks, printed photo textures, etc. 

The State-of-The-Art (SoTA) face PAD algorithms show good performance due to the large number of open-set databases available. However, they struggle somewhat with generalization to unseen capture devices or Presentation Attack Instruments (PAI). However, PAD systems for ID cards remain in their infancy, primarily due to the difficulty of obtaining sufficient bona fide samples \cite{tapia2024competitionpresentationattackdetection} and the lack of open-set databases. The ID cards contain sensitive information used to identify subjects; therefore, genuine bona fide samples cannot be published. This is why all publicly available datasets are used as bona fide samples of synthetic or simulated ID cards printed on Polyvinyl Chloride Plastic (PVC). The simulated bona fide samples (synthetic) are insufficient for research and for a production-ready PAD system.

The first competition on PAD on ID card was organized in the last International Joint Conference on Biometrics 2024 \cite{tapia2024competitionpresentationattackdetection}. The results showed that the submitted algorithms lacked generalization capabilities. The 2025 edition of this challenge \cite{tapia2025secondcompetitionpresentationattack} showed improvements in the generalization of the algorithms tested, with one team surpassing the baseline. However, it remains challenging because the baseline and the best submitted algorithm differed only slightly when evaluated across four countries.

One of the main drawbacks of PAD algorithms for ID cards is their poor generalization to unseen document types or country versions \cite{rocamora2024fewshotlearningexpandingid}. Each country has its own ID card template, and multiple versions may coexist during certain periods. Each template varies significantly in its background, typography, layout, and other elements. This heterogeneity affects the performance of PAD systems, making generalization difficult. In Sanchez et al. \cite{rocamora2024fewshotlearningexpandingid}, meta-learning techniques based on Few-shot learning were explored. It was suggested that at least $100$ bona fide ID card users and their respective attacks were needed to obtain good performance in a new type of ID. However, even $100$ bona fide users are difficult to obtain due to privacy concerns, and generating the corresponding attacks requires significant time and effort. Therefore, it is crucial to explore alternatives that could reduce this number.

Several open-set datasets are available on the SoTA. Some of the open-set datasets available are MIDV-500 \cite{MIDV-500}, MIDV-2020 \cite{MIDV2020AC}, MIDV-Holo \cite{midv-holo}, DLC-2021 \cite{dlc2021}, and SynIDPass \cite{SynID}, which usually contain a small number of synthetic users, printed on PVC, which is not enough to train a robust production-ready PAD system. 

Motivated by the previous challenge, we proposed a new method toward Zero-shot based on Few-shot learning to predict on ID cards from a new country using pretrained features extracted from a PAD representing a different country. The main contributions of this work are:

\begin{itemize}
    \item A Few-Shot approach is proposed to reduce the number of images needed to extend PAD to other countries.
    \item A comparison with a Zero-shot foundation model was explored to extend the generalization capabilities
    \item Our method was evaluated on private and open-set datasets using IDs from different countries.
\end{itemize}

The remainder of the article is organized as follows: Section~\ref{sec:related} summarizes related works on PAD and \textit{FSL}; Section~\ref{sec:method} describes the proposed method; Section~\ref{sec:experimentation} describes the experimentation performed to test the proposed method; finally, Section~\ref{sec:conclusion} discusses the results and future work.

\section{Related work}
\label{sec:related}

Few-shot learning \textit{(FSL)} has emerged as an effective learning method with considerable potential~\cite{song}. Despite recent creative efforts to tackle \textit{FSL} tasks, rapidly learning valid information from just a few or even zero samples remains a significant challenge.

Song et al.~\cite{song} propose and analyze, in detail, a survey of more than $200$ works that have raised the open challenge of \textit{FSL} in recent years, due to the restrictions and limited access to data imposed by privacy preservation issues.

The \textit{FSL} approaches usually work with two sets of samples: a \textit{query set} containing samples from unknown classes and a \textit{support set} containing samples from known classes. The \textit{FSL} approaches use the \textit{support set} as a source of information to infer the class of each sample of the query set. Depending on the constitution of the \textit{support set}, the nomenclature \textit{N-way-K-shot} is used where $N$ is the number of classes and $K$ the number of samples of each class in the \textit{support set}. When $K = 0$, it is called \textit{Zero-Shot} learning; otherwise, the term \textit{K-Shot} or \textit{Few-Shot} is used.

In the literature, we found two main approaches based on \textit{FSL} theory. One of them is the Relation Network \textit{(RL)} \cite{relation}, and the other is the Prototypical Network \textit{(PN)} \cite{prototypical}, which stands out for its simplicity.

Both networks are based on K-Nearest Neighbors \textit{(KNN)}, but with the advantage of being trainable end-to-end. The \textit{KNN} can be considered as an \textit{FSL} approach that computes the distance between the \textit{query sample} and all \textit{support samples}, returning the most common class among the $K$ closest support samples. The problem with \textit{KNN} is that its effectiveness depends on the discriminatory power of the embeddings used to compute the distance. The embeddings usually come from a previously trained backbone, and they are traditionally trained to optimize a Multi-Layer Perceptron \textit{(MLP)} classifier, which produces an embedding space that may not be optimal for computing distances between samples of unknown class. In addition, traditional \textit{KNN} requires lots of samples of each class to work correctly. 

Both \textit{RL} and \textit{PN} try to solve these issues by providing an end-to-end trainable \textit{KNN}. The \textit{RL} replaces the \textit{MLP} for a module that learns a distance function to optimize the discriminatory power of both the embeddings and the distance function. 

On the other hand, \textit{PN} replaces the \textit{MLP} with a one-hot similarity score layer with a \textit{softmax} as a loss function to optimize the distance between \textit{query samples} and a \textit{prototype} of each class. This prototype can be computed as the average embedding of the class. The ability of both networks to be trained end-to-end, along with training that uses a different subset of classes per batch, enables them to perform well on unseen classes.

Sanchez et al. \cite{rocamora2024fewshotlearningexpandingid} used the Prototypical Network to improve generalization with a few samples. The Prototypical Network is a metric-based \textit{FS} learner trained to assign samples to their class prototypes. This prototype is computed by averaging the embeddings of a few samples of the class. This average, along with the training process in which the prototypes are changed in each batch, allows the network to generalize to new prototypes. In that work, additional samples of new ID card types were incrementally added to the training set to establish a benchmark for the \textit{PN} ability to learn from a few samples. That work concludes that at least $100$ distinct bona fide users and their respective attacks were required to achieve a good performance.

In addition, architectures based on foundation models such as (Contrastive Language–Image Pre-training) \textit{CLIP} \cite{clip} have revolutionized the \textit{Zero-Shot} scene. The \textit{CLIP}  is a family of vision-language models trained on large-scale image-text pairs to learn joint representations of visual and textual data. The \textit{CLIP-B/16} model uses a $16\times16$ patch size, making it a medium-sized model that balances performance and efficiency. It is trained on 400 million image-text pairs from the internet and excels at zero-shot classification, image retrieval, and transfer learning. In contrast, \textit{CLIP-L/14} uses a \textit{ViT-L/14} (Vision Transformer Large, $14\times14$ patches) architecture, which is significantly larger and more powerful. With over 1 billion parameters, it utilizes more complex visual and semantic relationships, achieving SoTA performance. 

The \textit{CLIP} is a Vision Transformer architecture \textit{(ViT)} \cite{vit} based network that has been trained to compare the embedding generated by an image and its text description. This enables \textit{Zero-Shot} by computing \textit{Cosine Similarity} (CS) between the textual class descriptions and the images of those classes.

\section{Proposed method: Few-Shot learning approach}
\label{sec:method}
In previous work, such as Sanchez et al. \cite{rocamora2024fewshotlearningexpandingid}, the authors aim to expand into new countries or domains with reduced data use. In that work, the authors indicate that $100$ bona fide samples and their attacks are needed to obtain competitive results. These $100$ samples, in an ID card problem, aren't a small number. 

Our approach is based on the premise that identifying the ID card type is easier than the PAD itself, so we assume that there is an ID card type classifier that selects the appropriate \textit{support set}. For example, we tested an approach that is trained on different types of ID cards, driving licenses, and passports. This classifier uses the first half of an EfficientNetV2-b0 \cite{effv2} backbone to generate embeddings. Those embeddings are then classified using a KNN that performs cosine distance to the embedding of the document template or specimen. This approach achieves an average accuracy of 98.5\% across more than 150 different documents and 99.8\% on the ID cards tested in this work. That classifier was trained with both real and synthetic samples. Although the classifier is out-of-scope for this work, it is essential to note that we could train one using only synthetic samples or specimens.

\begin{figure}[t]
    \centering
    \begin{subfigure}{0.48\linewidth}
        \centering
        \includegraphics[scale=0.6]{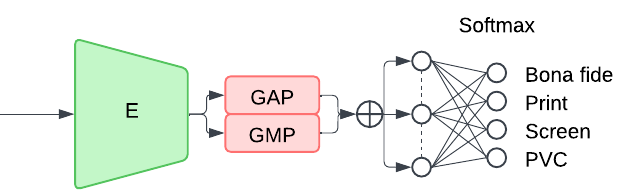}
        \caption{}
        \label{fig:diagram-baseline}
    \end{subfigure}
    \hfill
    \begin{subfigure}{0.48\linewidth}
        \centering
        \includegraphics[scale=0.5]{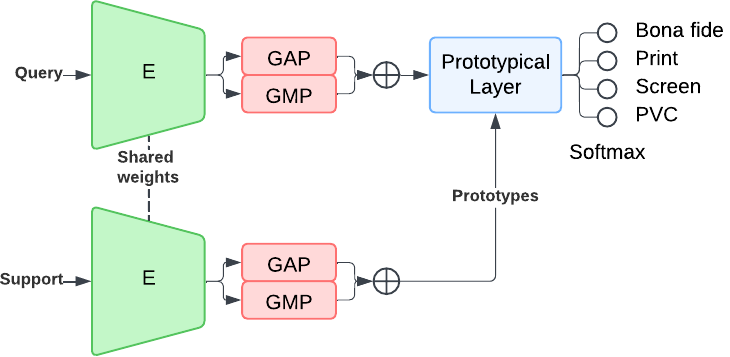}
        \caption{}
        \label{fig:diagram-proto-train}
    \end{subfigure}

    \vspace{1em}
    \begin{subfigure}{0.48\linewidth}
        \centering
        \includegraphics[scale=0.5]{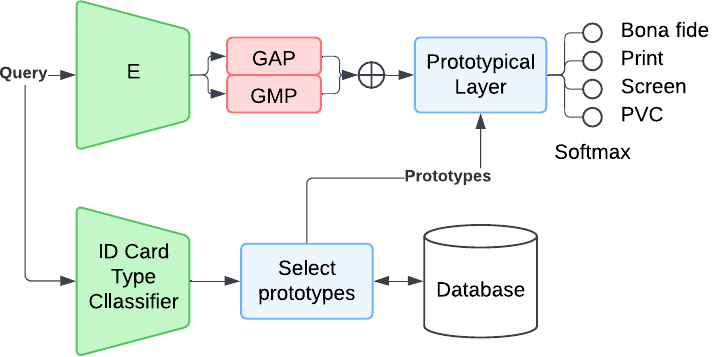}
        \caption{}
        \label{fig:diagram-proto-test}
    \end{subfigure}

    \caption{Diagrams for baseline method (a), Prototypical Network train stage (b), and Prototypical Network inference stage (c).}
    \label{fig:diagram}
\end{figure}

A diagram of our method is shown in Figure \ref{fig:diagram}. First of all, \ref{fig:diagram-baseline}, the baseline is shown, with a backbone (marked with an $E$ in Figure), where an embedding is extracted by concatenating the Global Average Pooling \textit{(GAP)} and Global Max Pooling \textit{(GMP)} of the backbone output. At the head, a hidden layer and a Softmax Classifier Layer with four classes are present: bona fide, print, screen, and PVC, respectively. 

To build our system, we modify the baseline by replacing the hidden layer with a \textit{Prototypical Layer}. According to the \textit{N-Way-K-Shot} nomenclature, our proposed method is a \textit{4-Way-K-Shot} because we have $4$ classes. The $K$ is defined in preliminary experiments, which couldn't be exposed in this work due to length limitations. We defined \textit{a support set} of $4$ samples per class. Prototypes were built by averaging the embeddings of $4$ samples per class of the target ID card type.

In the experiments, fewer samples (one, two, or three) showed high dependence on the selected samples, and with 4 samples, this is the lowest number at which the results are stable. We selected the lowest because we are in a scenario where bona fide samples are limited. In that case, our method is a \textit{4-Way-4-Shot} method that uses only $16$ samples: $4$ bona fide samples and their respective attacks. In our method, the model never uses any samples of the target ID card type during training; \textit{it only uses them during inference to build the prototypes}. Unlike \cite{rocamora2024fewshotlearningexpandingid}, where the samples are used to fine-tune the model and to build the prototypes together.

\vspace{-0.3cm}

\subsection{Training Stage}
Our approach proposes improving the number of images used in the previously proposed method by leveraging \textit{Few-Shot} training. We use episodic training. An episodic task is a single, self-contained training task with its own start and end, such as a game match or a specific classification problem. Train a model to recognize these particular classes using only the support set. To test, evaluate the model on new examples (query set) of those same classes, then create new episodes with different classes and examples, and repeat the process, allowing the model to learn how to learn across many other, small-scale challenges. 

In our case, each batch comprises the same number of samples per class (bona fide, print, screen, and PVC) but includes only a single ID card type. This is similar to classic Few-shot episodic training; however, \textit{instead of changing the classes in each episode, we change the domain}. This enhances the Prototypical Network's ability to transfer across domains simply by changing the \textit{prototypes}. The Prototypical head uses the \textit{Euclidean distance} to compute the \textit{logits} function. The Figure~\ref{fig:diagram-proto-train} shows the diagram of the PN training process.

\subsection{Inference Stage}
During inference, we rely on an ID card type classifier to select the prototypes for the current sample. To work with a new ID card type, we will need samples of each class to build the prototypes: $4$ bona fide and their respective attacks. These prototypes can be precomputed and stored in a database to be fetched on demand. Figure~\ref{fig:diagram-proto-test} shows the inference process.

\subsection{Configuration and hyperparameters}
We used a backbone based on \textit{EfficientNet-V2-b0} as a SoTA \cite{rocamora2024fewshotlearningexpandingid}. Therefore, we had an embedding of size $1\times2,560$, as shown in Figure~\ref{fig:diagram}. 

We used the unaligned card crop, resized to $224\times224$ as input. A batch size of $84$ \textit{query samples} ($21$ samples per class) and $16$ \textit{support samples} ($4$ samples per class). We used $4$ class-output classifiers during training and validation with the \textit{softmax} activation. However, we used the bona fide class as a binary classifier during inference. Regarding hyperparameters, we used \textit{AdamW} as the optimizer with a learning rate of $5e-4$. We trained for up to $100$ pseudo-epochs of $200$ steps, with early stopping over the last $10$ epochs.

\subsection{Databases and Metrics}
The databases used to evaluate and test the proposed methodology are described below. Two databases were used: a private database to enable extensibility to production systems and a public database to replicate the results of this work. For this purpose, to evaluate the performance of the proposed PAD method, the metrics described in ISO/IEC 30107-3\footnote{\url{https://www.iso.org/standard/79520.html}} are to be used. The Equal Error Rate (EER), Bona fide Presentation Classification Error Rate (BPCER), and Attack Presentation Classification Error Rate (APCER). To further evaluate performance under fixed security constraints, the metric BPCER at fixed APCER (BPCER$_{AP}$) is employed. 

The BPCER measures the proportion of bona fide presentations that are wrongly classified as attacks. The following expression defines it:

\begin{equation}
    BPCER = \frac{1}{N_{BF}} \sum_{i=1}^{N_{BF}} RES_i,
\end{equation}

where $N_{BF}$ is the number of bona fide presentation samples, and $RES_i = 1$ if the sample is incorrectly flagged as an attack.

The APCER is defined as:
\begin{equation}
    APCER_{PAIS} = 1 - \frac{1}{N_{PAIS}} \sum_{i=1}^{N_{PAIS}} RES_i,
\end{equation}

where $N_{PAIS}$ represents the total number of attack images for a given PAIS, and $RES_i = 1$ if the $i-th$ image is correctly identified as an attack, or $0$ if it is misclassified as bona fide.

\subsubsection{Private Dataset}
A private database with different ID card types and bona fide, paper print, screen replay, and PVC print classes is used to test our approach. Table~\ref{tab:num_samples} shows the different ID card types and the number of users of each one. As seen there, the Chilean ID card type \textit{CHL-2} has significantly more users than the others. 

\begin{figure}[ht]
    \centering
    \begin{subfigure}{\textwidth}
        \centering
        \includegraphics[scale=0.15]{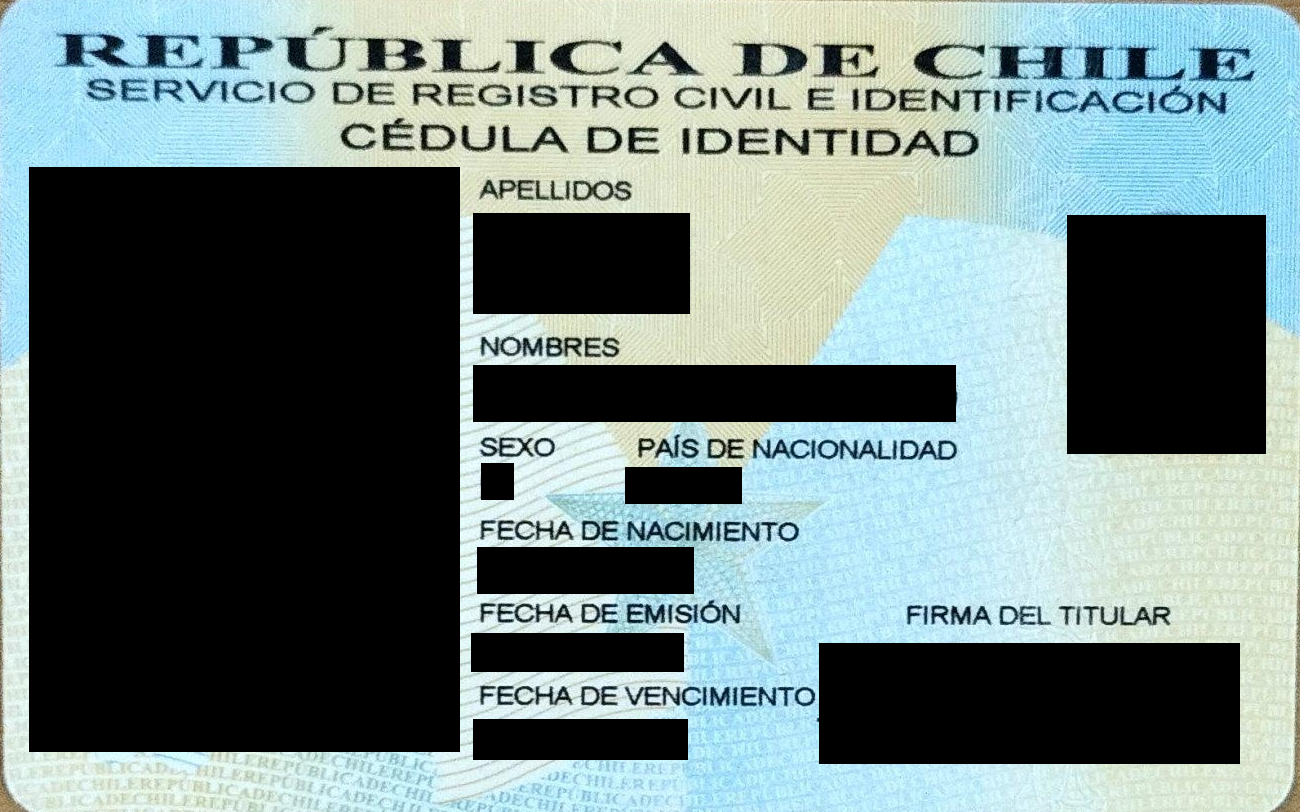}%
        \includegraphics[scale=0.225]{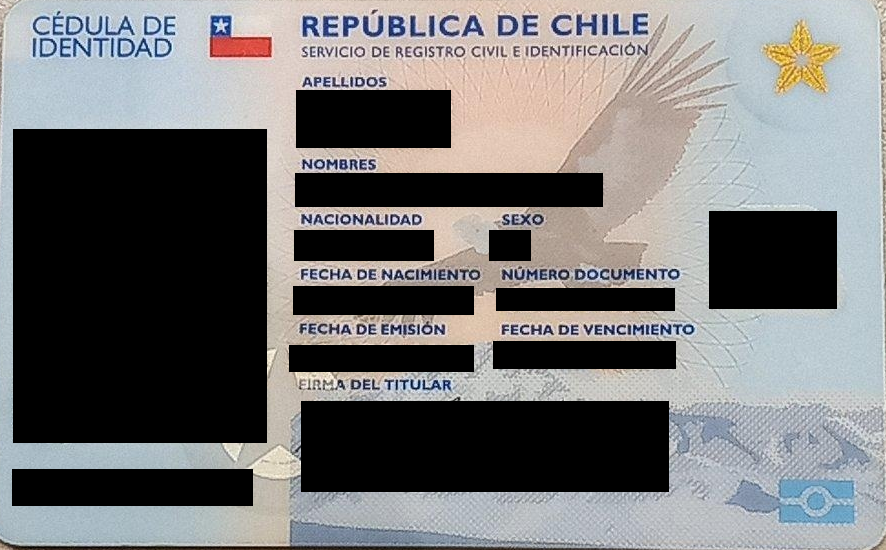}%
        \includegraphics[scale=0.08]{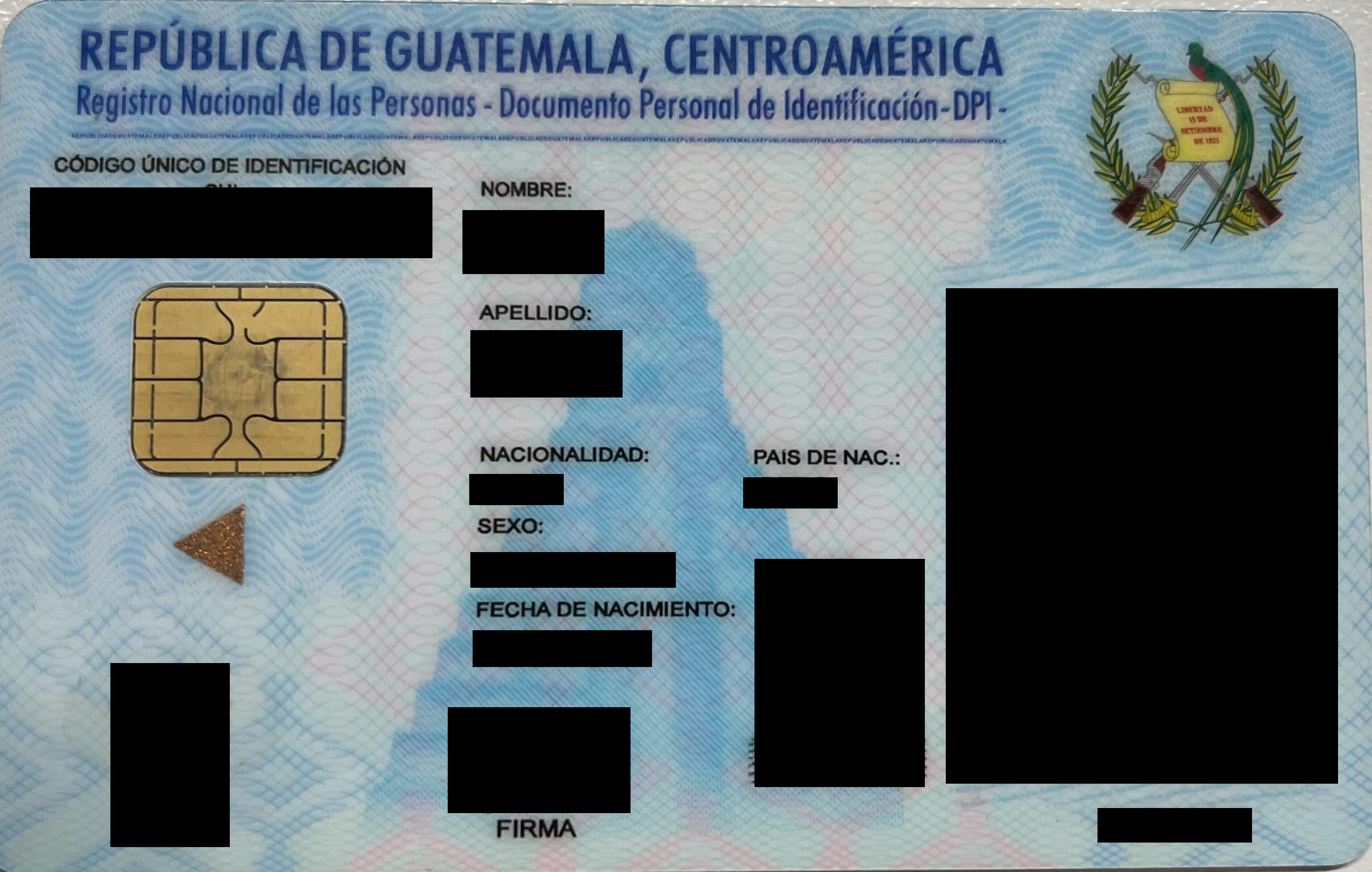}%
        \caption{Example images from CHL-1, CHL-2 and Guatemala}
        \label{fig:examples-CHL1-CHL2-GTM1}
    \end{subfigure}
    \begin{subfigure}{\textwidth}
        \centering
        \includegraphics[scale=0.11]{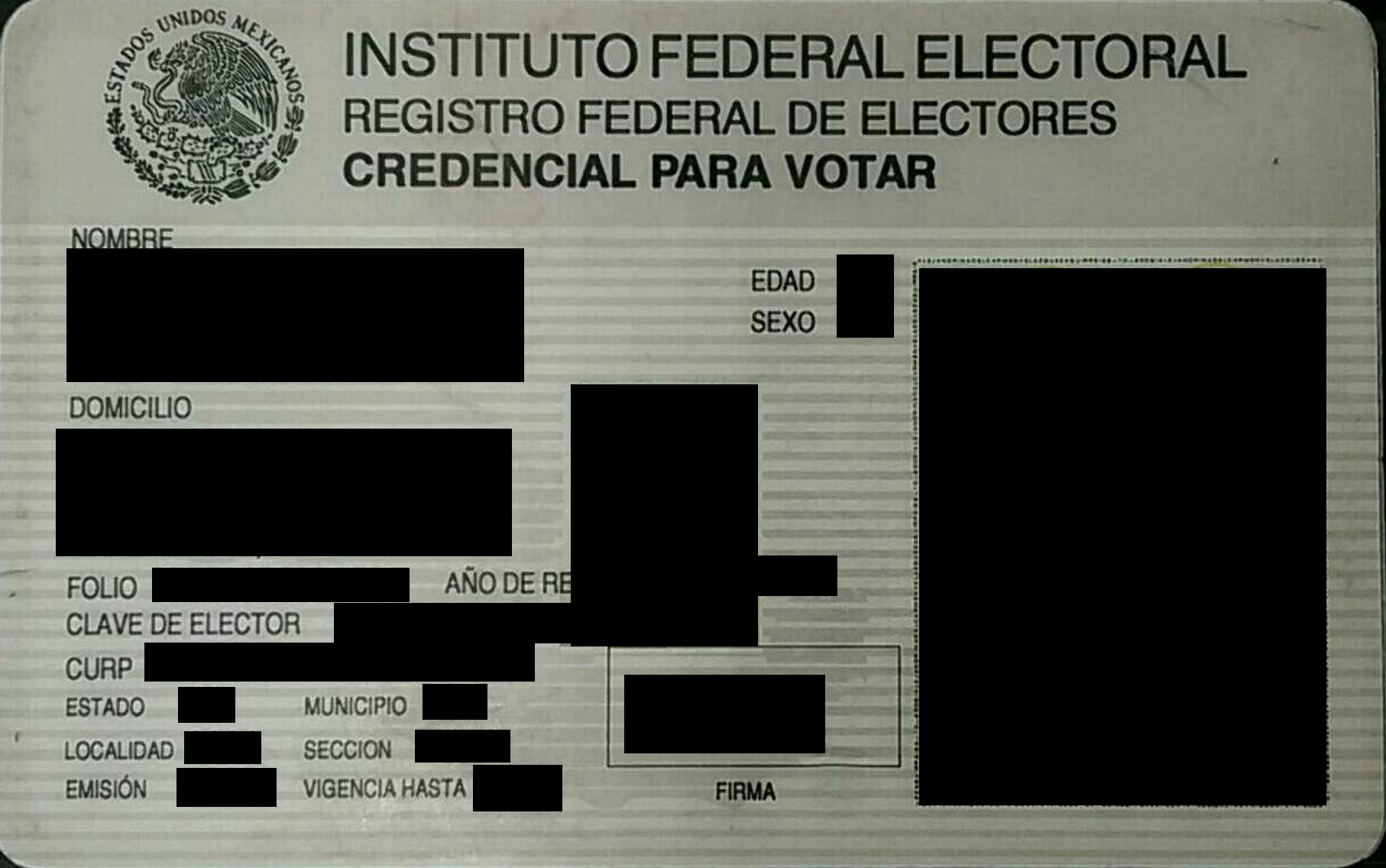}%
        \includegraphics[scale=0.16]{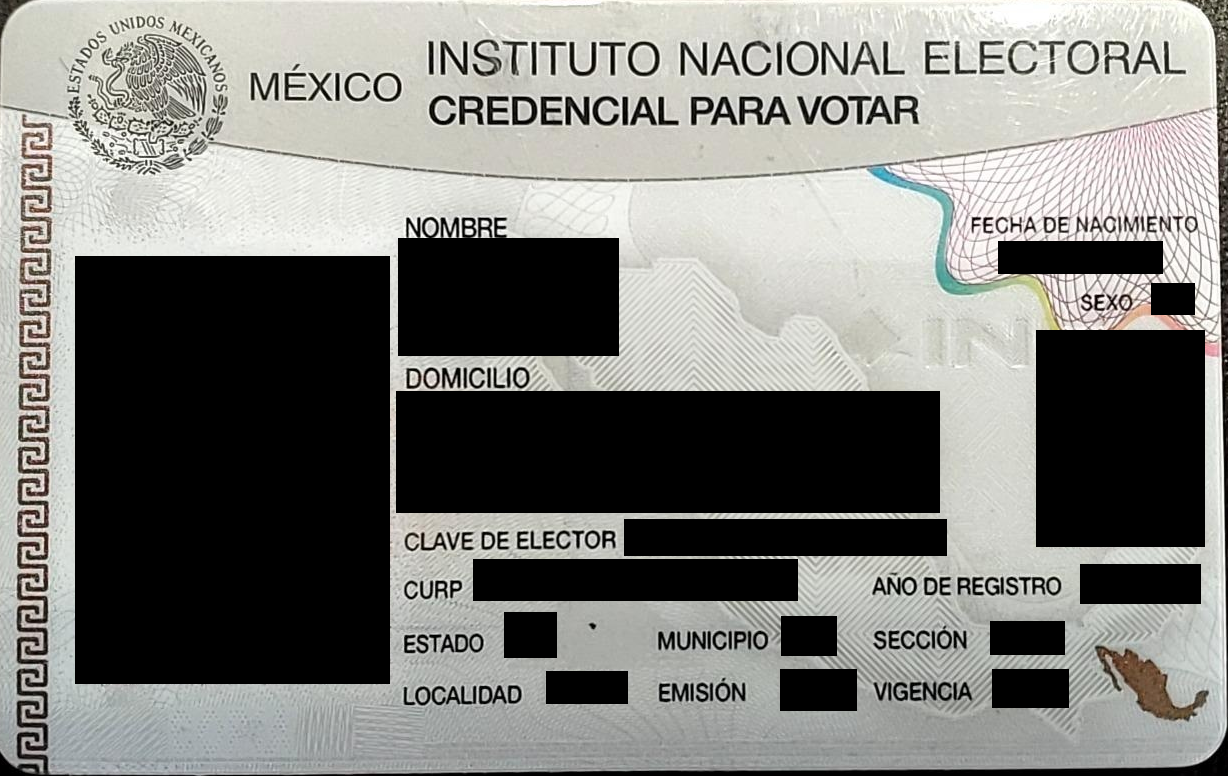}%
        \includegraphics[scale=0.13]{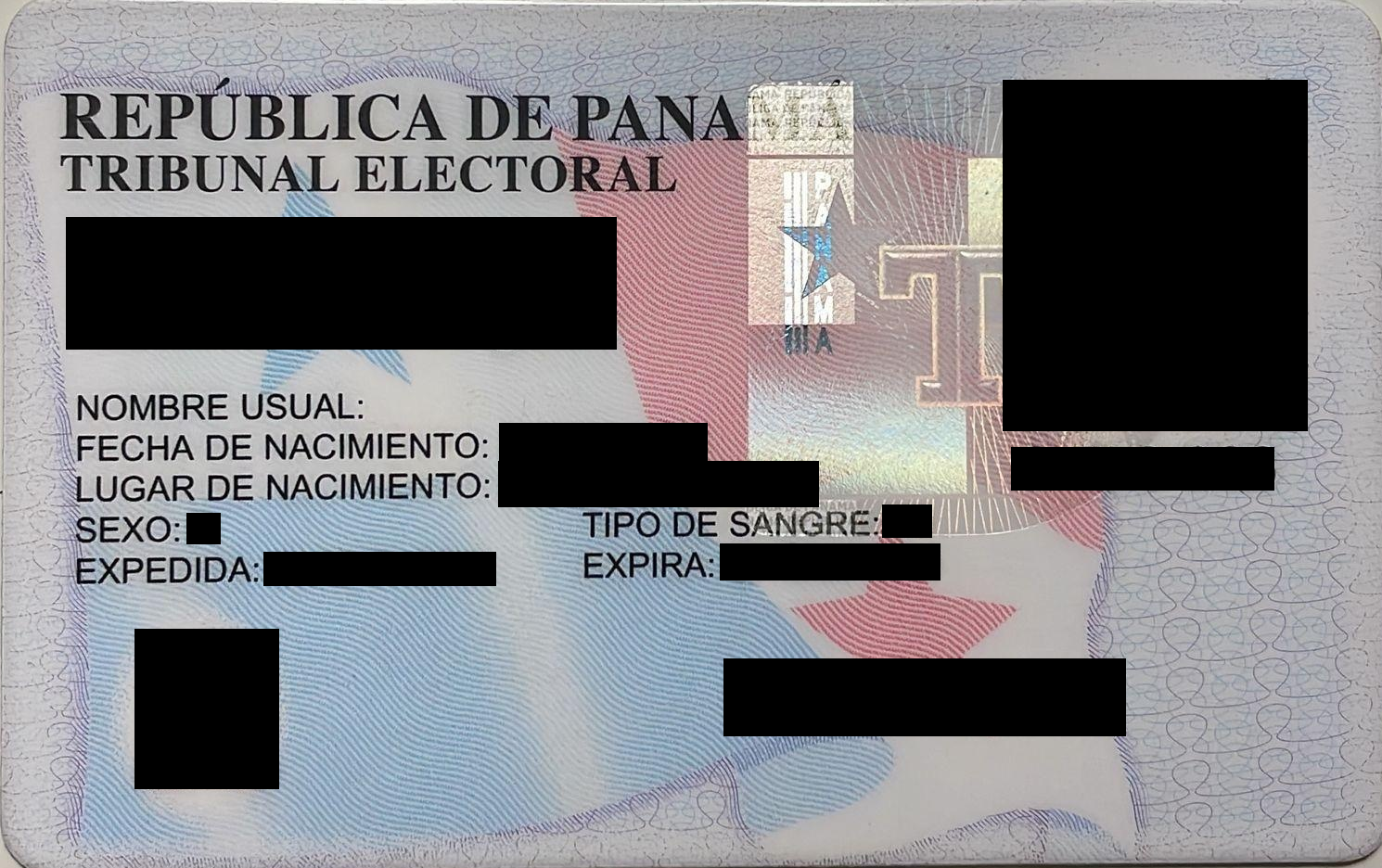}%
        \caption{Example images from Mexico-1, Mexico-2 and Panama.}
        \label{fig:examples-MEX1-MEX2-PAN1}
    \end{subfigure}
    \caption{Example of images from a private dataset.}
    \label{fig:examples}
\end{figure}

\subsubsection{Public dataset DLC-2021}
To ensure reproducibility of the proposed experiments, we have replicated the experiments using a publicly available dataset. The chosen dataset is DLC-2021 \cite{dlc2021}, which contains synthetic samples of different ID card types and passports with their corresponding screen and print attacks. We used only the ID card samples from Albania \textit{(ALB)}, Spain \textit{(ESP)}, Estonia \textit{(EST)}, Finland \textit{(FIN)} and Slovakia \textit{(SVK)}.  

\begin{table}[]
\label{tb:dataset}
\scriptsize
\centering
\caption{Number of samples per ID card type.}
\label{tab:num_samples}
\begin{tabular}{@{}rccccccc@{}}
\toprule
\multicolumn{1}{l}{} & \textbf{PAN} & \textbf{COL} & \textbf{CHL-1} & \textbf{CHL-2} & \textbf{MEX-1} & \textbf{MEX-2} & \textbf{GTM} \\ \midrule
\textbf{Bona fide}   & 1,006         & 80           & 696            & 13,000          & 353            & 660            & 1,001 \\
\textbf{Print}       & 1,006         & 80           & 696            & 13,000          & 353            & 660            & 1,001 \\
\textbf{Screen}      & 1,006         & 80           & 696            & 13,000          & 353            & 660            & 1,001 \\
\textbf{PVC}         & 506          & 80           & 259            & 6,500           & 353            & 160            & 501 \\ \bottomrule
\end{tabular}%
\end{table}

\section{Experiments and Results}
\label{sec:experimentation}

\subsection{Experiment 0: CLIP Zero-Shot}
Firstly, we evaluated \textit{CLIP Zero-Shot} capabilities to compare them with our approach. We performed some evaluations using only the ID card for most samples, which is \textit{CHL-2}, to adjust the text prompts and determine which \textit{CLIP} backbone yields better results. The best performing configuration was used for comparison with our approach in both experiments.

We tested the backbone \textit{CLIP-B16} and \textit{CLIP-L14} from \textit{OpenAI}, following these simple text prompts:
\begin{itemize}
\item \textit{Bona fide: ``a photo of an ID card"}
\item \textit{Print: ``a printed photo of an ID card"}
\item \textit{Screen: "a photo of an ID card shown on a screen device",}
\item \textit{PVC: "a photo of an ID card printed on a PVC card"},
\end{itemize}

Because a PVC-printed ID card might be mistaken for the class-printed one, we also tested without that text class to see whether the results improved. Results are shown in the Table~\ref{tab:clip_test}. We observe improvements in both $B-16$ and $L-14$ backbones when only $3$ text classes are used. The backbone $B-16$ also outperforms the backbone $L-14$ on $3$ classes. Therefore, we used $B-16$ with only $3$ text classes as the \textit{Zero-Shot} approach.

\begin{table}[H]
\scriptsize
\centering
\caption{Experiment 0: CLIP test with CHL-2 in private dataset.}
\label{tab:clip_test}
\begin{tabular}{@{}rcccc@{}}
\toprule
\multicolumn{1}{l}{Model/Classes} & \textbf{EER\%} & \textbf{BPCER10} & \textbf{BPCER20} & \textbf{BPCER100} \\ \midrule
\textbf{CLIP-B16 / 4C}        & 30.53       & 66.66           & 81.63          & 95.24           \\
\textbf{CLIP-L14 / 4C}        & 30.86       & 65.31           & 78.44          & 94.21            \\
\textbf{CLIP-B16 / 3C}        & \textbf{20.71}       & \textbf{41.52}           & \textbf{60.14}          & 88.06           \\
\textbf{CLIP-L14/ 3C}        & 27.07       & 51.33           & 65.63          & \textbf{86.33}          \\
\hline
\end{tabular}
\end{table}

\subsection{Experiment 1: Single domain training}
In this experiment, we used the most common ID card type, \textit{CHL-2}, with most samples to train both the baseline and prototypical networks. In this case, the episodic training consists of keeping the same classes and the same ID card type, but changing the samples of the support set in each episode. The goal of this experiment is to assess the generalization capabilities to unseen ID Card types when only a single ID card type is available during training. This is the most challenging case.

The results shown in Table~\ref{tab:exp1} indicate that the baseline does not generalize to unseen ID card types. However, the Prototypical model shows better generalization. The best-performing ID card type for the baseline is \textit{PAN}, which achieves an EER of 12.7\% and a \textit{BPCER10} of 17.5\% while the prototypical reaches 8.44\% and 7.06\%, respectively. 
On average, the baseline achieves an EER of 26.56\% and a \textit{BPCER10} of 45.69\%, while the prototypical achieves 9.07\% and 10.15\%, respectively, which is substantially better. 

Comparing both to \textit{Zero-Shot CLIP-B16}, we can observe that \textit{CLIP} performs better than the baseline but worse than the prototypical approach.

\begin{table}[H]
\scriptsize
\centering
\caption{Experiment 1 on the private dataset in (\%).}
\label{tab:exp1}
\begin{tabular}{@{}rcccc@{}}
\toprule
\multicolumn{1}{l}{} & \textbf{EER\%} & \textbf{BPCER10} & \textbf{BPCER20} & \textbf{BPCER100} \\ \midrule
\multicolumn{1}{l}{} & \multicolumn{4}{c}{\textbf{Baseline}}                                  \\ \midrule
\textbf{PAN}        & 12.70       & 17.50           & 31.11          & 56.96           \\
\textbf{COL}        & 38.54      & 66.25          & 75.00             & 92.5            \\
\textbf{CHL-1}        & 21.35      & 36.64          & 40.80           & 71.84           \\
\textbf{MEX-1}        & 56.80      & 100.00         & 100.00         & 100.00          \\
\textbf{MEX-2}        & 56.80      & 100.00         & 100.00         & 100.00          \\
\textbf{GTM}        & 16.73      & 33.87          & 48.55          & 65.83           \\ \midrule
\textbf{AVG}         & 26.56      & 45.69         & 53.79          & 75.96           \\
\textbf{STD}         & 17.62      & 31.80          & 28.28          & 16.65           \\ \bottomrule
\toprule
\multicolumn{1}{l}{} & \multicolumn{4}{c}{\textbf{Zero-Shot CLIP-B16}}                                  \\ \midrule
\textbf{PAN}        & 18.31      & 27.53          & 42.94          & 66.30           \\
\textbf{COL}        & 22.29      & 38.75          & 47.5           & 78.75            \\
\textbf{CHL-1}        & 14.79      & 22.27          & 36.21          & 65.80           \\
\textbf{MEX-1}        & 8.17       & 7.08           & 13.31          & 34.28          \\
\textbf{MEX-2}        & 17.37      & 28.33          & 44.70          & 77.58          \\
\textbf{GTM}        & 11.04      & 12.19          & 23.88          & 65.03           \\ \midrule
\textbf{AVG}         & 15.33      & 22.69          & 34.76          & 64.62           \\
\textbf{STD}         & 5.13       & 11.55          & 13.48          & 16.08           \\ \bottomrule
\toprule
\multicolumn{1}{l}{} & \multicolumn{4}{c}{\textbf{Prototypical Network}}                                  \\
\midrule
\textbf{PAN}        & 8.44       & 7.06           & 14.21          & 36.18           \\
\textbf{COL}        & 1.25       & 0.00           & 0.00           & 1.25           \\
\textbf{CHL-1}        & 12.93      & 19.68          & 36.49          & 72.99           \\
\textbf{MEX-1}        & 10.76      & 12.75          & 25.21          & 58.92           \\
\textbf{MEX-2}        & 11.37      & 12.42          & 25.76          & 63.18           \\
\textbf{GTM}        & 9.69       & 8.99           & 24.28          & 54.95           \\ \midrule
\textbf{AVG}         & 9.07       & 10.15          & 20.99          & 47.91           \\
\textbf{STD}         & 4.12       & 6.58           & 12.48          & 25.88           \\ \bottomrule
\end{tabular}
\end{table}

\subsection{Experiment 2: Multiple domain training}
In this experiment, we performed Leave-One-Out (LOO) cross-validation on \textit{CHL-2}, \textit{PAN}, \textit{COL}, and \textit{MEX-1}, in which we trained both the baseline and the prototypical model using $N-1$ ID card types and held out the remaining ID card type for testing. In this case, episodic training keeps the same classes while varying the ID Card type. The goal of this experiment is to assess the generalization capabilities when multiple ID card types are available during training.

Observing the results in Table~\ref{tab:exp2}, we can see that the advantage of the prototypical is reduced. In this experiment, the best-performing ID card type for the baseline is \textit{PAN}, which achieves an EER of 3.06\% and a \textit{BPCER10} of 0.6\%, surpassing the prototypical, which achieves only 5.67\% and 1.49\%, respectively. However, on average, the prototypical outperforms the baseline: the baseline achieves an EER of 11.98\% and a \textit{BPCER10} of 16.95\%, while the prototypical achieves 7.41\% and 5.88\% respectively.

Comparing with the \textit{Zero-Shot CLIP-B16}, we can observe that both the baseline and the prototypical now perform better. The only exception is in \textit{MEX-1}, where \textit{CLIP} outperforms both of them.

\begin{table}
\scriptsize
\centering
\caption{Experiment 2 on the private dataset in (\%).}
\label{tab:exp2}
\begin{tabular}{@{}rcccc@{}}
\toprule
\multicolumn{1}{l}{} & \textbf{EER\%} & \textbf{BPCER10} & \textbf{BPCER20} & \textbf{BPCER100} \\ \midrule
\multicolumn{1}{l}{} & \multicolumn{4}{c}{\textbf{Baseline}}                                  \\ \midrule
\textbf{PAN}        & 3.06       & 0.60           & 1.69           & 6.96            \\
\textbf{COL}        & 14.17      & 20.00          & 36.25          & 66.25           \\
\textbf{CHL-2}        & 11.95      & 14.91          & 29.34          & 63.37           \\
\textbf{MEX-1}        & 18.74      & 32.29          & 45.04          & 81.87           \\ \midrule
\textbf{AVG}         & 11.98      & 16.95          & 28.08          & 54.61           \\
\textbf{STD}         & 6.58       & 13.12          & 18.73          & 32.79           \\ \bottomrule
\toprule
\multicolumn{1}{l}{} & \multicolumn{4}{c}{\textbf{Zero-Shot CLIP-B16}}                                  \\ \midrule
\textbf{PAN}        & 18.31      & 27.53          & 42.94          & 66.30           \\
\textbf{COL}        & 22.29      & 38.75          & 47.5           & 78.75            \\
\textbf{CHL2}        & 20.71      & 41.52          & 60.14          & 88.06           \\
\textbf{MEX1}        & 8.17       & 7.08           & 13.31          & 34.28          \\
\midrule
\textbf{AVG}         & 17.37      & 28.72          & 40.97          & 66.85           \\
\textbf{STD}         & 6.35       & 15.64          & 19.82          & 23.47           \\ \bottomrule
\toprule
\multicolumn{1}{l}{} & \multicolumn{4}{c}{\textbf{Prototypical Network}}                                  \\ \midrule
\textbf{PAN}        & 5.67       & 1.49           & 6.46           & 25.65           \\
\textbf{COL}        & 3.54       & 0.00           & 0.00           & 12.50           \\
\textbf{CHL-2}        & 11.86      & 16.34          & 39.31          & 81.10           \\
\textbf{MEX-1}        & 8.55       & 5.67           & 19.55          & 63.46           \\ \midrule
\textbf{AVG}         & 7.41       & 5.88           & 16.33          & 45.68           \\
\textbf{STD}         & 3.61       & 7.38           & 17.34          & 32.00           \\ \bottomrule
\end{tabular}
\end{table}
\vspace{-0.3cm}

\subsection{Experiments with the public dataset - DLC-2021}
Now, the same experiments are replicated using the \textit{DLC-2021} dataset. 
\subsubsection{Experiment 0}
\label{sec:exp_0_PD}

For \textit{CLIP}, we tested the same backbones with \textit{ESP} ID cards using only $ 3$-class text prompts because, in \textit{DLC-2021}, the bona fide is a PVC-printed ID card. We tested two sets of prompts: one with the bona fide prompt as the bona fide class, and another with the PVC prompt as the bona fide class. \\
The results in Table~\ref{tab:clip_test_dlc} show that using the bona fide prompt (bf-prompt) produces the worst results compared to using the bona fide PVC-prompt, especially in \textit{BPCER100}. The \textit{CLIP-B16} backbone is the best in \textit{BPCER100}; however, in all cases, performance is quite low compared to the results obtained on the private dataset. This is because the bona fide samples of \textit{DLC-2021} are PVC-printed ID cards, which are often confused with the Print class text prompt. We chose the \textit{CLIP-B16} backbone using the PVC prompt to compare with our approach.

\begin{table}[H]
\scriptsize
\centering
\caption{CLIP tests (\%) with ESP ID cards in DLC-2021.}
\label{tab:clip_test_dlc}
\begin{tabular}{@{}rcccc@{}}
\toprule
\multicolumn{1}{l}{} & \textbf{EER\%} & \textbf{BPCER10} & \textbf{BPCER20} & \textbf{BPCER100} \\ \midrule
\textbf{B16-bf-Prompt}        & 49.37       & 93.39           & 97.82          & 100.00           \\
\textbf{L14-bf-Prompt}        & 38.05       & 89.99           & 95.96          & 99.42          \\ 
\textbf{B16-bf-PVC}             & 41.81       & 81.33           & 90.64          & 96.92           \\
\textbf{L14-bf-PVC}             & 42.67       & 89.16           & 96.47          & 99.17          \\ \bottomrule
\end{tabular}
\end{table}

\subsubsection{Experiment 1}
\label{sec:exp_1_PD}
Replicating experiment 1 using the \textit{DLC-2021} database, the results are similar to those obtained with our private dataset, as shown in Table~\ref{tab:exp1_dlc}. In Experiment 1, the baseline does not generalize to unseen ID cards, with an average EER of 27.8\%, while the prototypical achieves an average EER of 10.53\%. \\

\begin{table}[H]
\scriptsize
\centering
\caption{Experiment 1 on the DLC-2021 dataset in (\%).}
\label{tab:exp1_dlc}
\begin{tabular}{@{}rcccc@{}}
\toprule
\multicolumn{1}{l}{} & \textbf{EER\%} & \textbf{BPCER10} & \textbf{BPCER20} & \textbf{BPCER100} \\ \midrule
\multicolumn{1}{l}{} & \multicolumn{4}{c}{\textbf{Baseline}}                                  \\ \midrule
\textbf{ALB}        & 26.91       & 53.43           & 70.32           & 85.92            \\
\textbf{EST}        & 21.34       & 35.65           & 44.32            & 68.44          \\
\textbf{FIN}        & 27.12      & 53.22          & 64.15          & 82.57           \\
\textbf{SVK}        & 35.84      & 71.78          & 83.96          & 94.13           \\ \midrule
\textbf{AVG}         & 27.8      & 53.77          & 65.69          & 82.77           \\
\textbf{STD}         & 5.99       & 15.17          & 16.48          & 10.71           \\ \bottomrule
\toprule
\multicolumn{1}{l}{} & \multicolumn{4}{c}{\textbf{Zero-Shot CLIP-B16}}                                  \\ \midrule
\textbf{ALB}        & 38.13      & 76.65          & 88.54          & 98.35           \\
\textbf{EST}        & 41.96      & 87.18          & 94.14          & 98.90           \\
\textbf{FIN}        & 39.71      & 75.70          & 83.51          & 93.63           \\
\textbf{SVK}        & 36.66      & 84.15          & 91.59          & 98.90           \\ \midrule
\textbf{AVG}        & 39.12      & 80.92          & 89.45          & 97.45           \\
\textbf{STD}        & 2.27       & 5.63           & 4.57           & 2.56           \\ \bottomrule
\toprule
\multicolumn{1}{l}{} & \multicolumn{4}{c}{\textbf{Prototypical Network}}                                  \\ \midrule
\textbf{ALB}        & 13.4       & 18.18           & 33.6           & 77.85           \\
\textbf{EST}        & 5.98      & 0.92          & 8.24          & 35.59           \\
\textbf{FIN}        & 9.99       & 9.74           & 25.8         & 51.34           \\
\textbf{SVK}        & 12.73       & 17.35           & 31.77          & 89.7           \\ \midrule
\textbf{AVG}         & 10.53       & 11.55           & 24.85         & 63.62           \\
\textbf{STD}         & 3.37       & 8.04           & 11.56         & 24.62           \\ \bottomrule
\end{tabular}
\end{table}
\vspace{-0.3cm}

\subsubsection{Experiment 2}
\label{sec:exp_2_PD}
In Experiment 2 and Table~\ref{tab:exp2_dlc}, the advantage of the Prototypical is reduced; the average EER is 10.27\% for the baseline and 9.12\% for the prototypical. In this case, the advantage of the Prototypical is lost when setting the threshold to \textit{BPCER100}, where the baseline achieves an average of 45.15\% while the prototypical reaches 61.6\%.

Comparing results with \textit{CLIP-B16}, we can observe that \textit{CLIP} does not perform well with \textit{DLC-2021}. Both the baseline and the prototypical models significantly outperform \textit{CLIP}, which achieves an average EER of 39.65\%. 

\begin{table}[H]
\scriptsize
\centering
\caption{Experiment 2 on the DLC-2021 dataset in (\%).}
\label{tab:exp2_dlc}
\begin{tabular}{@{}rcccc@{}}
\toprule
\multicolumn{1}{l}{} & \textbf{EER\%} & \textbf{BPCER10} & \textbf{BPCER20} & \textbf{BPCER100} \\ \midrule
\multicolumn{1}{l}{} & \multicolumn{4}{c}{\textbf{Baseline}}                                  \\ \midrule
\textbf{ALB}        & 7.97       & 6.92           & 12.67           & 47.43            \\
\textbf{ESP}        & 5.29       & 1.78           & 5.47            & 29.09           \\
\textbf{EST}        & 2.95       & 0.31           & 0.67            & 7.88           \\
\textbf{FIN}        & 20.10      & 36.41          & 49.34          & 82.26           \\
\textbf{SVK}        & 15.05      & 21.16          & 33.52          & 59.11           \\ \midrule
\textbf{AVG}         & 10.27      & 13.32          & 20.34          & 45.15           \\
\textbf{STD}         & 7.13       & 15.32          & 20.5          & 28.39           \\ \bottomrule
\toprule
\multicolumn{1}{l}{} & \multicolumn{4}{c}{\textbf{Zero-Shot CLIP-B16}}                                  \\ \midrule
\textbf{ALB}        & 38.13      & 76.65          & 88.54          & 98.35           \\
\textbf{ESP}        & 41.81      & 81.33          & 90.64          & 96.62           \\
\textbf{EST}        & 41.96      & 87.18          & 94.14          & 98.90           \\
\textbf{FIN}        & 39.71      & 75.70          & 83.51          & 93.63           \\
\textbf{SVK}        & 36.66      & 84.15          & 91.59          & 98.90           \\ \midrule
\textbf{AVG}        & 39.65      & 81.00          & 89.68          & 97.34           \\
\textbf{STD}        & 2.31       & 4.88           & 3.99           & 2.23           \\ \bottomrule
\toprule
\multicolumn{1}{l}{} & \multicolumn{4}{c}{\textbf{Prototypical Network}}                                  \\ \midrule
\textbf{ALB}        & 10.28       & 10.34           & 16.95           & 84.33           \\
\textbf{ESP}        & 4.55       & 0.43           & 3.26           & 74.05           \\
\textbf{EST}        & 5.37      & 2.93          & 6.04          & 31.99           \\
\textbf{FIN}        & 8.87       & 7.18           & 16.43          & 42.91           \\
\textbf{SVK}        & 16.54       & 24.84           & 41.76          & 74.72           \\ \midrule
\textbf{AVG}         & 9.12       & 9.15           & 16.89          & 61.6           \\
\textbf{STD}         & 4.78       & 9.57           & 15.18          & 22.75           \\ \bottomrule
\end{tabular}
\end{table}

\section{Conclusion and future work}
\label{sec:conclusion}
We have shown that it is possible to generalize to unseen ID card types using only \textit{four bona fide users and their respective attacks}. The experiments show that our approach outperforms the baseline considerably when only one ID card type is available during training. However, in Experiment 2, where more ID card types are available during training, this advantage is reduced. Even though the Prototypical still outperforms the baseline. The need for an ID card type classifier can sometimes be problematic. Therefore, we recommend using our Prototypical approach when only a few ID card samples (four) are available during training; when there are sufficient ID card types, the baseline's generalization capabilities are adequate, so there is no clear advantage to using a more complex architecture such as the Prototypical.

One of the main drawbacks of this approach is the need for bona fide and attack samples. To address this issue, we will use an anomaly detection approach in which we train the prototypical model to determine whether a sample is similar to the bona fide prototype. This could reduce the samples required to only the bona fides of the target ID card. Furthermore, we will investigate the possibility of using only specimen samples as a prototype, so we will not even need any bona fide users.

\subsubsection{Acknowledgements} 
This research has been partially funded by Facephi, R\&D department, and the European Union's Horizon research and innovation program under grant agreements 101121280 (EINSTEIN) and the German Federal Ministry of Education and Research, and the Hessian Ministry of Higher Education, Research, Science, and the Arts, which jointly support the National Research Center for Applied Cybersecurity ATHENE.

%
%
%
\bibliographystyle{splncs04}
\bibliography{DAS/References}

\end{document}